\documentclass{article}

\usepackage{PRIMEarxiv}

\usepackage[utf8]{inputenc} 
\usepackage[T1]{fontenc}    
\usepackage{hyperref}       
\usepackage{url}            
\usepackage{booktabs}       
\usepackage{amsfonts}       
\usepackage{nicefrac}       
\usepackage{microtype}      
\usepackage{fancyhdr}       
\usepackage{graphicx}       

\graphicspath{{media/}}     

\fancypagestyle{firstpage}{
  \fancyhf{} 
  \fancyhead[C]{\small\textcolor{gray}{This work has been submitted to the IEEE for possible publication. \\Copyright may be transferred without notice, after which this version may no longer be accessible.}}
}

\usepackage{subcaption}
\usepackage{booktabs}
\usepackage{float}
\usepackage{tikz}
\usepackage{array}
\usepackage{comment}
\usepackage{diagbox}
\usepackage{amsmath}

\DeclareMathOperator*{\argmax}{arg\,max}
\DeclareMathOperator*{\argmin}{arg\,min}
\newcommand{\ours}{{\textit{AdaptIn}}} 

\title{Context-Aware Input Orchestration for Video Inpainting
}

\author{
  Hoyoung Kim \\
  Department of Computer Science \\
  Stony Brook University \\
  NY, United States\\
   \And
  Azimbek Khudoyberdiev \quad Seonghwan Jeong \quad Jihoon Ryoo \thanks{Correspondence to: Jihoon Ryoo (jihoon.ryoo@sunykorea.ac.kr)} \\
  Department of Computer Science \\
  The State University of New York, Korea \\
  Incheon, South Korea\\
}

\begin{document}
\maketitle
\thispagestyle{firstpage}

\begin{abstract}
Traditional neural network-driven inpainting methods struggle to deliver high-quality results within the constraints of mobile device processing power and memory. Our research introduces an innovative approach to optimize memory usage by altering the composition of input data. Typically, video inpainting relies on a predetermined set of input frames, such as neighboring and reference frames, often limited to five-frame sets. Our focus is to examine how varying the proportion of these input frames impacts the quality of the inpainted video. By dynamically adjusting the input frame composition based on optical flow and changes of the mask, we have observed an improvement in various contents including rapid visual context changes.

\end{abstract}


\begin{figure}[h]
\begin{center}
\centerline{\includegraphics[width=\linewidth]{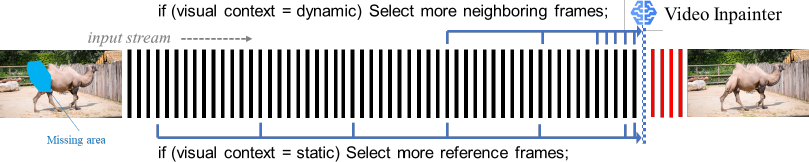}}
\caption{Conceptual Description of Input Configuration for Video Inpainting. In the scenario of inpainting frames streamed, we adapt the input composition based on the dynamics of the visual context to efficiently utilize memory while preserving quality.}
\label{fig:main_concept}
\end{center}
\end{figure}

\section{Introduction}

Video inpainting, a technique aimed at filling in missing or damaged regions within a video, has garnered significant attention in recent years due to its potential applications in video restoration, editing, and enhancement. In the realm of video inpainting, addressing the challenges about cost of memory remains a critical concern. Due to these challenges, while there are many applications of image inpainting on mobile devices, video editing remains difficult to implement. Specifically, handling lengthy videos or high-resolution content proves challenging due to the substantial memory requirements, necessitating the exploration of more efficient memory utilization strategies.



Motivated by these challenges, our research sets out to explore novel perspectives in video inpainting. We begin by hypothesizing that the importance of reference frames in input frames is more pronounced when the contextual information within a video remains static. Conversely, in dynamic contexts, the influence of neighboring frames is expected to surpass that of reference frames. These hypotheses serve as the foundation for our investigation.

In this paper, we aim to validate these hypotheses by employing optical flow completion in mask regions. By observing and sharing insights into the influence of input configuration based on the optical flow, we address the correlation between contextual dynamics and inpainting quality. Specifically, we apply optical flow completion to mask areas, allowing us to adaptively adjust the composition of input frames, thereby improving the inpainting quality in fast contextual change. The concept is depicted in Figure \ref{fig:main_concept}.

To summarize the key points this paper aims to deliver:
\begin{itemize}
  \item It discusses the correlation between contextual dynamics and inpainting quality in memory-constrained environments.
  \item It proposes a pipeline for structuring the input by considering the dynamics and further emphasizes the importance of accounting for these dynamics when designing models.
\end{itemize}

We are going to observe that the input frame composition has influence on the video inpainting quality in Section \ref{subsec:psnr_videos}. In Section \ref{subsec:psnr_flow&mask}, we will explore the correlation of this phenomenon with visual context including optical flow and mask changes. To utilize the observation, in Section \ref{sec:system}, we introduce a pipeline, {\ours}, involving simple but effective input configuration. In Section \ref{sec:experiment}, we will investigate how {\ours} can improve inpainting quality by compose input frames differently based on the visual context.

Through the empirical validation of our hypotheses, this research contributes to the advancement of video inpainting methodologies, offering insights into memory-efficient strategies and paving the way for improved inpainting quality in both static and dynamic contexts.
\section{Related Works}


Neural network-based video inpainting has witnessed remarkable advancements, with researchers exploring diverse approaches. In this literature review, we categorize existing methodologies into flow-based, Transformer-based, and fusion approaches.

\subsection{Flow-based Approaches}
VINet \cite{kim2019deep} introduces a deep neural network architecture for fast video inpainting. The framework leverages an image-based encoder-decoder model to collect and refine information from neighboring frames and synthesize unknown regions. Temporal consistency is enforced through ConvLSTM layers and a flow estimator.

FGVC \cite{Gao-ECCV-FGVC} proposes a novel flow-edge based video completion method that leverages motion edges to achieve sharper and more accurate results. This approach first extracts and completes motion edges, which are then used to guide the flow completion process.

\subsection{Transformer-based Approaches}
STTN \cite{yan2020sttn} employs self-attention mechanisms within the transformer network to learn joint spatial and temporal relationships between pixels, enabling consistent and accurate inpainting.

FuseFormer \cite{Liu_2021_FuseFormer} proposes a Transformer model leveraging fine-grained feature fusion. The core idea lies in two operations: Soft Split, which splits feature maps into overlapping patches, and Soft Composition, which aggregates information from these patches. These operations facilitate the capture of fine-grained details and their effective integration within the Transformer architecture.

\subsection{Fusion (Transformer + Optical Flow) Approaches}
E$^{2}$FGVI \cite{liCvpr22vInpainting} utilizes flow information to guide a Transformer model. It consists of three main modules - a flow completion module to estimate and complete optical flow fields, a feature propagation module that uses the completed flows to guide the propagation of features between frames, and a content hallucination module comprising temporal-spacial transformer blocks.

ProPainter \cite{zhou2023propainter} introduces dual-domain propagation and a mask-guided sparse Transformer. Dual-domain propagation combines image and feature warping, enabling more effective and reliable global correspondence exploitation. The mask-guided sparse Transformer achieves high efficiency by discarding unnecessary tokens, significantly reducing computational cost while maintaining performance.

The previous studies have demonstrated impressive results in the inpainting task. However, their methods often fail to account for visual dynamics, resulting in outputs that may appear incongruous in either static or highly dynamic sequences. In Section \ref{sec:observation}, we discuss the variability in quality of existing models across static and dynamic video content. In Section \ref{sec:system}, we will propose a method to address these limitations and enhance the overall quality and consistency of inpainting results across diverse video dynamics.
\section{Influence of Composition in Input Frames}
\label{sec:observation}

\begin{figure}[h]
\begin{center}
\centerline{\includegraphics[width=0.6\columnwidth]{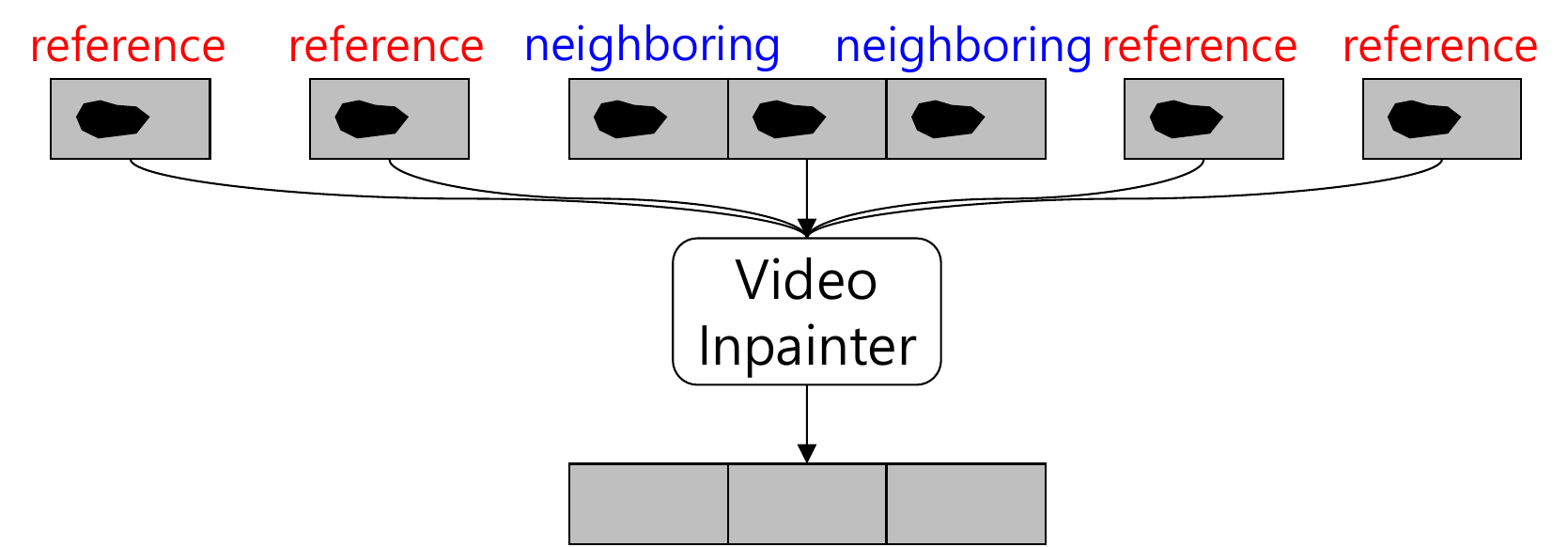}}
\caption{Example of reference frames and neighboring frames as input for video inpainting. Neighboring frames refer to the input frames that are either the target frame to be inpainted or frames adjacent to the target frame. Reference frames are those frames that are temporally distant from the target frames.}
\label{fig:example_ref_neigh}
\end{center}
\end{figure}

Existing inpainting models proposed thus far take a fixed number of neighboring frames and distant reference frames as input, as shown in Figure \ref{fig:example_ref_neigh}. 
The higher the number of input frames, the better the quality, but it is not feasible when available memory is limited. However, the frames influencing inpainting results may vary depending on the context of the video. In this section, we will examine how the inpainting quality differs based on the ratio of reference frames in the input.

We generate moving mask with random shape on every frame. 
The dataset is \textit{DAVIS 2017} \cite{pont2017davis}, where the videos have a relatively short length of around 2 seconds. The reference frames are extracted at intervals of 10 frames.
The number of input frames is 8.
The models used for testing in this section include the Transformer model STTN \cite{yan2020sttn} and the fusion approach ProPainter \cite{zhou2023propainter}.

\subsection{Inpainting Quality and Composition of Input Frames across Videos}
\label{subsec:psnr_videos}

\begin{figure}[h]
    \centering
    \includegraphics[width=\linewidth, trim={4.7cm 0 4.7cm 0}]{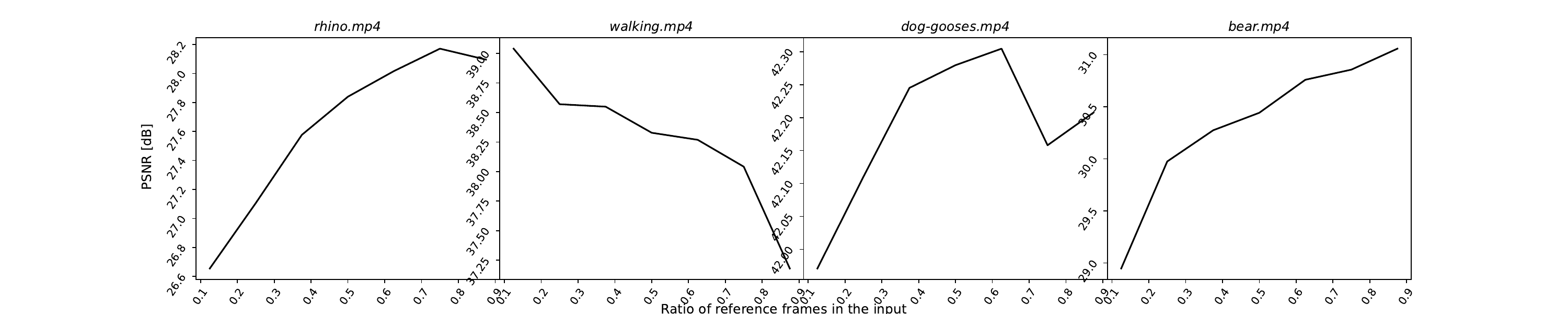}
    \caption{PSNR over the Ratio of Reference Frames in Input Frames across Different Videos. The x-axis represents the ratio of reference frames within the input frames. As x increases, the proportion of reference frames becomes higher, whereas a lower value indicates a higher proportion of neighboring frames. The PSNR patterns vary according to this ratio for each video. In the cases of \textit{rhino.mp4}and \textit{bear.mp4}, the PSNR increased as the proportion of reference frames increased. This can be interpreted as the visual context of these two videos being slow. Conversely, for \textit{walking.mp4}, the quality improved as the proportion of neighboring frames increased. This suggests that the visual context is fast and thus more influenced by the surrounding frames.}
    \label{fig:psnr_videos}
\end{figure}

We initially conducted experiments to observe how the PSNR quality varies for each video based on the composition ratio of input frames. PSNR (Peak Signal-to-Noise Ratio) is a metric to measure the similarity to the original \cite{psnrssim2010}. As presented in Figure \ref{fig:psnr_videos}, the patterns observed varied depending on whether the video was static or dynamic. While patterns can differ between videos, even for the same video, distinct patterns may emerge based on the location of the mask. It can be inferred that the visual context around the mask location significantly influences the inpainting quality. Therefore, we applied the optical flow completion adopted in ProPainter to the mask area and investigated the correlation between optical flow, input frame composition, and inpainting quality.

\subsection{Inpainting Quality and Input
Frame Composition across Optical Flow on Mask}
\label{subsec:psnr_flow&mask}

We used the pre-trained recurrent optical flow completion same as one in ProPainter. After completing optical flow in mask area, we normalized optical flow based on the mask size because we wanted to mitigate the influence of the mask size. To describe this with mathematical equation, the optical flow on mask is $(u, v)_{mask}=(\frac{dx}{dt}, \frac{dy}{dt})_{mask}$. The normalized amount of the optical flow based on the mask size can be presented as $\sqrt{u^2+v^2}/mask\_size$.

We also quantified the change or movement of the mask. If we denote the mask at time $t$ as $M_t$, then the change of the mask at time $t$ can be expressed as $\sum_{k=1}^{mn} |M_{k, t} - M_{k, t-1}|$, where $k$ is a position of each pixel and the frame dimension is $m \times n$.

To quantify the variation in PSNR induced by input frame composition, optical flow and mask changes, we used a Signed Maximum Change Rate in PSNR as a metric. Where a PSNR score depending on a ratio of reference frames in input, $r$, is ${psnr}_{r}$, and a set of PSNR is $P=\{{psnr}_{r}|r \in [0.125,0.25,...,0.875]\}$, the metric used here is

\begin{equation}
SIGN(\argmax_{r}{P} - \argmin_{r}{P}) \times \frac{\max{P} - \min{P}}{\max{P}}.
\end{equation}

In this metric, a large negative value indicates a lower proportion of reference frames, implying a greater influence of neighboring frames. Conversely, a large positive value indicates a higher influence of reference frames.

\begin{figure}[h]
    \centering
    \begin{subfigure}[tb]{.45\textwidth}
        \includegraphics[width=\columnwidth]{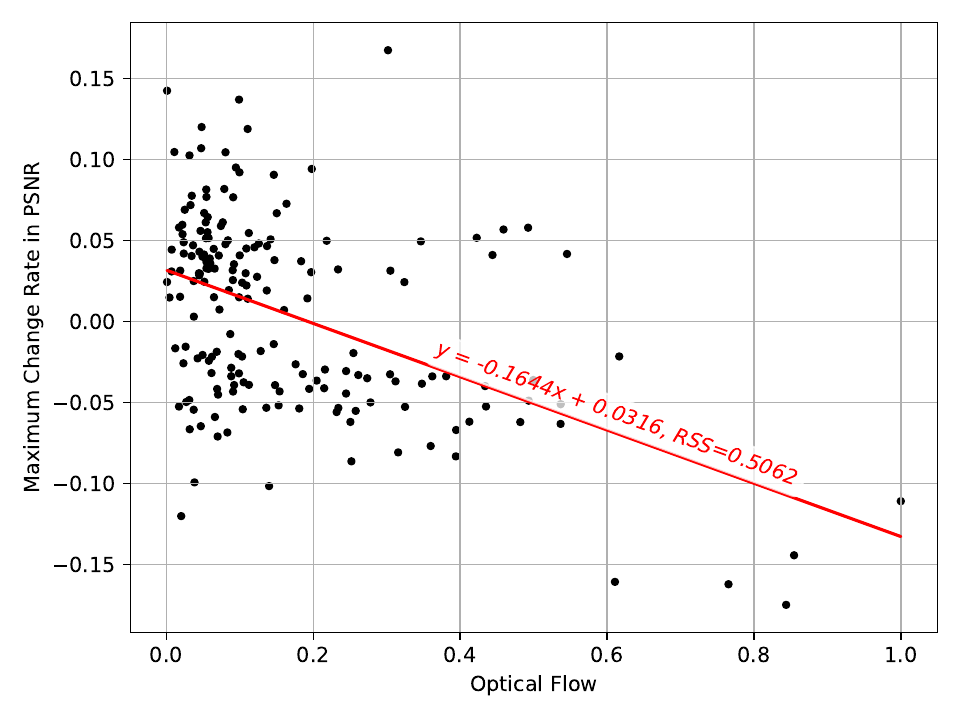}
        \caption{}
        \label{subfig:pro_dyn_flow}
    \end{subfigure}
    \begin{subfigure}[tb]{.45\textwidth}
        \includegraphics[width=\columnwidth]{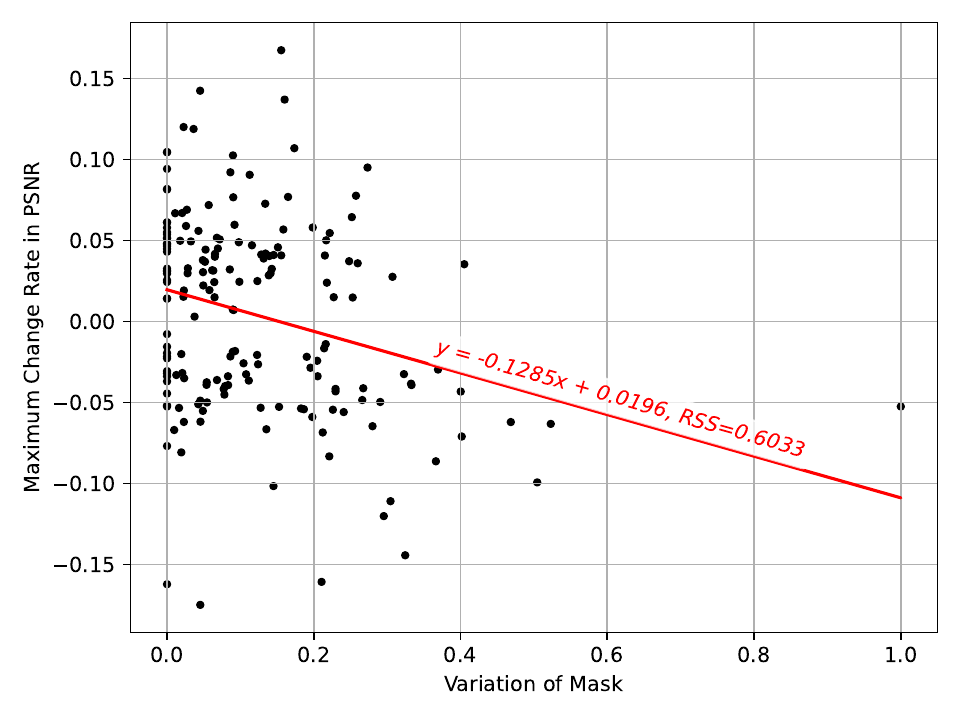}
        \caption{}
        \label{subfig:pro_dyn_mask}
    \end{subfigure}
    \caption{The Maximum Change Rates in PSNR across Optical Flow (a) and Change of Mask (b). The used model here is ProPainter. The y-axis represents the maximum change rate of PSNR according to the input frame composition, as shown in Figure \ref{fig:psnr_videos}. A positive value indicates that the PSNR is at its maximum when the proportion of reference frames is high, while a negative value indicates that the PSNR is at its maximum when the proportion of neighboring frames is high. As the optical flow and mask variation increase, the tendency shifts towards the negative direction. This indicates that the influence of neighboring frames becomes higher as the visual context becomes faster.}
    \label{fig:psnr_custom_metric}
\end{figure}

\subsubsection{Comparison of the Influence of Optical Flow and Mask Variation}

When the flow is significant and the change in the mask area, which is the region to be inpainted, is substantial, adjacent neighboring frames may have a greater influence on achieving better inpainting compared to reference frames.

Figure \ref{fig:psnr_custom_metric} demonstrates this tendency. The red line is the first order polynomial fit $f(x)$ into the data using least squared method \cite{bjorck1990least}. Both of Figure \ref{subfig:pro_dyn_flow} and Figure \ref{subfig:pro_dyn_mask} shows that, as the optical flow and the changes of mask become high, the changes in PSNR values becomes negative, it means neighboring frames have more influence for better inpainting quality. RSS is Residual Sum of Squares, $\sum_i^n (y_i - f(x_i))$, which smaller figure is better fit and lower error. This can be a clue how well the data follows the tendency. The RSS of \ref{subfig:pro_dyn_flow} is 0.5062 and the one of \ref{subfig:pro_dyn_mask} is 0.6033. Optical flow has stronger impact on inpainting quality rather than the change of mask.

\subsubsection{Comparison of the Effect of Flow on Inpainters based on Flow and Non-flow}

\begin{figure}[h]
\begin{center}
\centerline{\includegraphics[width=0.6\columnwidth]{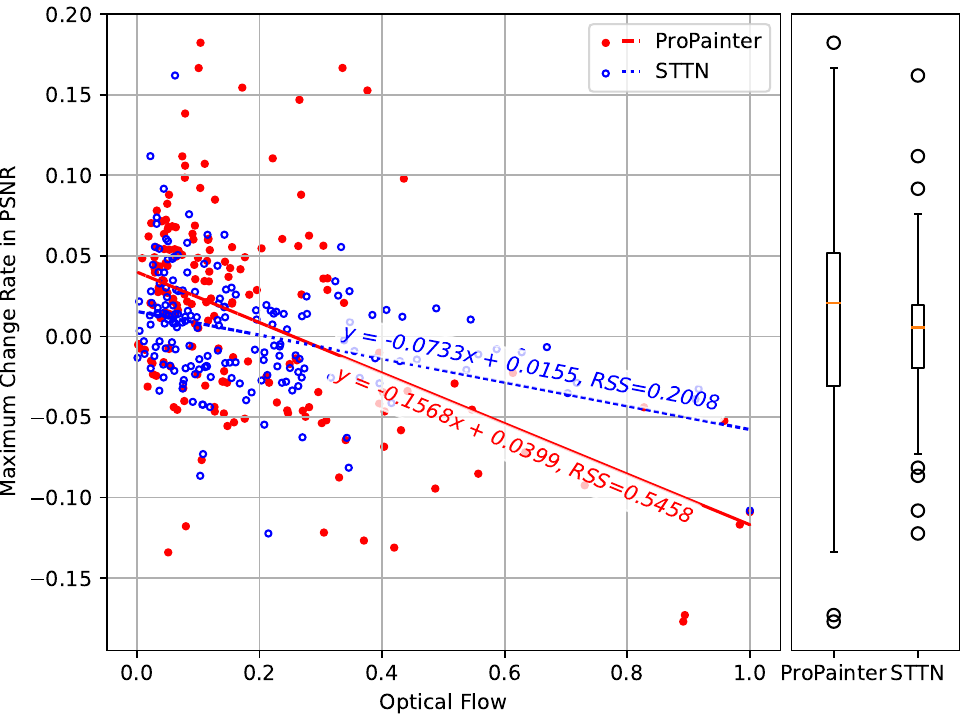}}
\caption{Comparison of PSNR Change Rates and their Distributions between the Non-flow-based Inpainter, STTN, and the Flow-guided Inpainter, ProPainter. The steeper the slope, the more it can be seen as being influenced by optical flow.}
\label{fig:compare}
\end{center}
\end{figure}

We compared the influence of optical flow on the inpainting quality for both non-flow-based approach, STTN, and flow-guided approach, ProPainter. As can be seen in Figure \ref{fig:compare}, ProPainter exhibits a heightened sensitivity to optical flow, as evidenced by the steeper slope of its line of best fit obtained through the least squared method. The box plot on the right clearly illustrates a more pronounced variation in PSNR for ProPainter as compared to STTN.

\begin{figure}[h]
\begin{center}
\centerline{\includegraphics[width=\columnwidth]{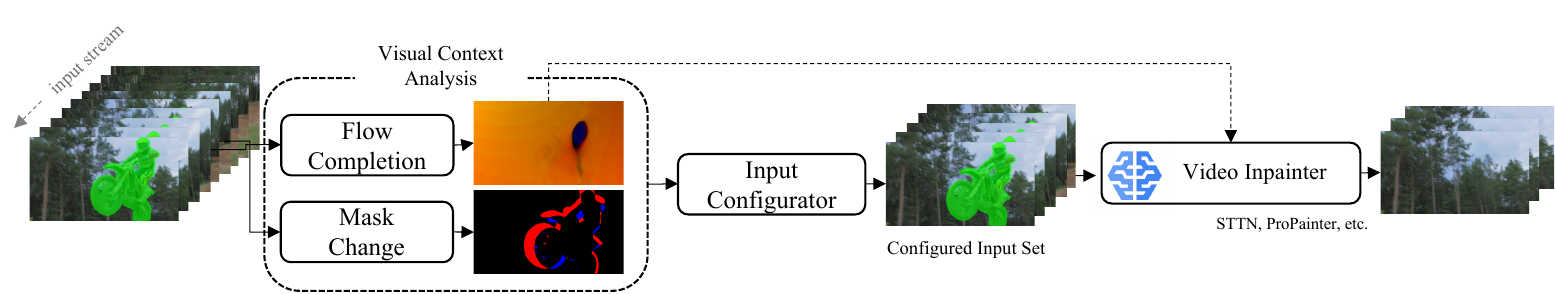}}
\caption{Pipeline of {\ours}.}
\label{fig:framework}
\end{center}
\end{figure}

\section{{\ours}}
\label{sec:system}

{\ours} is a pipeline devised based on the observation in Section \ref{sec:observation}. This is specifically designed for scenarios where a mobile device, including AR head-mounted displays, is tasked with processing streamed frames in real-time. Given the limitations in memory and computational power on such devices, {\ours} dynamically selects input frames for a video inpainting model based on a comprehensive analysis of the visual context.

\subsection{Visual Context Analysis}

The core functionality of {\ours} is illustrated in Figure \ref{fig:framework}, wherein contextual dynamics is analyzed using parameters, mask changes and optical flow. 

\paragraph{Mask change analysis}
{\ours} calculates the difference between the latest, two consecutive masks by summing the absolute values of their pixel-wise discrepancies, as defined in Section \ref{subsec:psnr_flow&mask}. This value is then normalized using the same scale as the one illustrated in Figure \ref{subfig:pro_dyn_mask}.

\paragraph{Optical flow analysis}
We apply the methodology used to obtain optical flow in ProPainter. Initially, optical flow between streamed frames is computed using RAFT model \cite{teed2020raft}. And then, the pre-trained recurrent model of ProPainter, which it consists of the second-order deformable alignment \cite{Chan_2022_CVPR}, completes the flow in masked regions. As explained in Section \ref{subsec:psnr_flow&mask}, {\ours} analyzes visual dynamics within inpainting areas based on completed flow values in the masked regions, rather than considering the entire frame.

\subsection{Input Configurator}
\label{subsec:configurator}

The pipeline intelligently adjusts the inputs with a greater number of neighboring frames during periods of highly dynamic context. Conversely, when the context is relatively static, {\ours} configures inputs with fewer neighboring frames and a higher proportion of reference frames.

To combine the analysis of optical flow and mask variations, we utilized the slope of Figure \ref{fig:psnr_custom_metric}. The integration of optical flow ($x_{flow}$) and mask variation ($x_{mask}$) into a new variable $x_{comb}$ is achieved by defining it as weighted arithmetic mean $x_{comb}=\frac{m_{flow}x_{flow} + m_{mask}x_{mask}}{m_{flow}+m_{mask}}$, where the slopes $m_{flow}$ and $m_{mask}$ are extracted from the lines of linear regression in Figure \ref{subfig:pro_dyn_flow} and \ref{subfig:pro_dyn_mask}, respectively.

\begin{figure}[h]
\begin{center}
\centerline{\includegraphics[width=0.7\columnwidth]{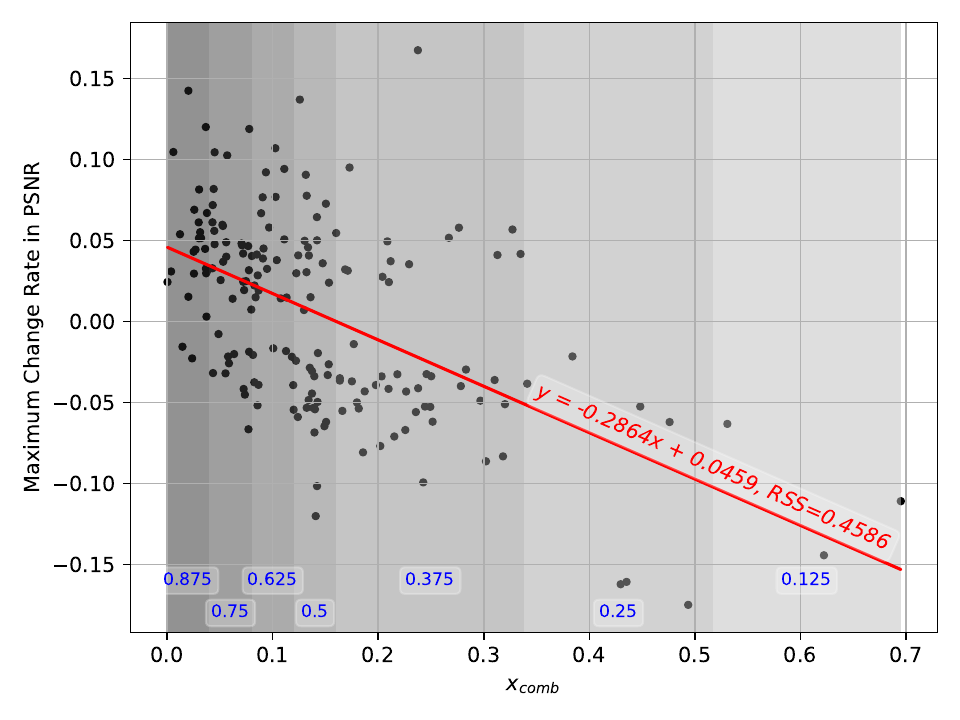}}
\caption{The Maximum Change Rate in PSNR according to $x_{comp}$ Combining both Optical Flow and Mask Variation. The blue numbers are the ratio of reference frames, $r_{ref}$, in each shaded region.}
\label{fig:combined}
\end{center}
\end{figure}

Figure \ref{fig:combined} shows the combined factor $x_{comp}$ results in a steeper slope of the red line for polynomial fit and lower RSS. Drawing insights from the linear trend depicted in Figure \ref{fig:combined}, we partitioned the range into seven segments. Within intervals exhibiting a PSNR variation of 0 or more, we evenly divided them into four ranges, assigning reference frame proportions equal or greater than the default. Conversely, for intervals with a PSNR variation below 0, we uniformly distributed them into three ranges, assigning reference frame proportions lower than the default. The blue text in the figure represents the reference frame proportions allocated in each shaded region.

\subsection{Video Inpainter}

The video inpainter block has existing neural-network-based inpainting models. These models fill in the pixels of the mask area based on the configured input set.

{\ours} re-uses the results obtained from the flow completion module. It is particularly beneficial for flow-guided inpainting models, such as ProPainter. By sharing the results from the flow completion module, {\ours} optimizes the inpainting process, offering enhanced efficiency and improved inpainting quality in scenarios with varying contextual dynamics.
\section{Experiment}
\label{sec:experiment}

\paragraph{Datasets}
For video completion, we create a random mask with random movement on each video. The proposed framework fills in the masked area which it completes the frames. To evaluate object removal task, \textit{DAVIS 2017} \cite{pont2017davis} and \textit{MOSE} \cite{MOSE} are utilized. Both datasets provide masks on objects. The videos are resized into $420\times240$. On setting for the evaluation, reference frames are extracted on every 10 frames. To sufficient comparison on different input configuration, videos with more than 120 frames are selected. Total video samples used for the evaluation are 45 from \textit{DAVIS 2017} and 171 from \textit{MOSE}.

\paragraph{Baseline}
We evaluated how {\ours} improves inpainting quality by applying different input frame compositions based on optical flow and mask change. The baseline configuration involved providing input data with an equal proportion (5:5 ratio) of the reference frames and the neighboring frames.

\paragraph{Memory Constraint}
While 'more inputs, higher quality' is obvious, the proposed framework targets on-device scenarios such as mobile phones or AR devices, which it may have memory limitation. When each input frame increases by one, memory usage increases by 662 MB on ProPainter and by 781 MB on STTN each time. We constrain the memory usage by feeding only 8 input frames. With 8 inputs, ProPainter consumed 5.365 GB and STTN consumed 6.310 GB of memory.

\subsection{Qualitative Evaluation}

\subsubsection{Video Completion}

\begin{table*}[htb]
\centering
\begin{tabular}{p{10mm}|l||rrrrrr}
\multicolumn{2}{c||}{\backslashbox{\begin{tabular}[c]{@{}c@{}}Input\\ Ratio\end{tabular}}{Dataset}} & \multicolumn{3}{c|}{\textit{DAVIS}} & \multicolumn{3}{c}{\textit{MOSE}} \\ \cline{1-8} 
\multicolumn{1}{c|}{$r_{ref}$} & \multicolumn{1}{c||}{$r_{nei}$} & $\Delta$PSNR$\uparrow$ & $\Delta$SSIM\textsuperscript{\dag}$\uparrow$ & \multicolumn{1}{l|}{$\Delta$VFID\textsuperscript{\ddag}$\downarrow$} & $\Delta$PSNR$\uparrow$ & $\Delta$SSIM\textsuperscript{\dag}$\uparrow$ & $\Delta$VFID\textsuperscript{\ddag}$\downarrow$ \\ \hline\hline
0.875 & 0.125 & 0.1958 & 103.3295 & \multicolumn{1}{r|}{-0.7076} & -0.1177 & -9.0709 & 6.0849 \\
0.75  & 0.25 & -0.2243 & -63.1119 & \multicolumn{1}{r|}{2.7133} & -0.1383 & -33.2395 & 1.1785 \\
0.625 & 0.325 & 0.0128 & 38.1968 & \multicolumn{1}{r|}{6.9960} & -0.0225 & 29.9844 & 0.6077 \\
0.5   & 0.5 & - & - & \multicolumn{1}{r|}{-} & - & - & - \\
0.325 & 0.625 & 0.1361 & 20.4277 & \multicolumn{1}{r|}{-4.5172} & 0.1067 & 49.9549 & -6.8513 \\
0.25  & 0.75 & 0.4786 & 63.3504 & \multicolumn{1}{r|}{-7.4989} & -0.0213 & -56.9942 & -15.5103 \\
0.125 & 0.875 & 3.2773 & 898.8341 & \multicolumn{1}{r|}{-25.5346} & 0.2729 & 172.7542 & -13.6241 \\ \hline
\multicolumn{2}{c||}{Overall} & 0.1092 & 30.4394 & \multicolumn{1}{r|}{-0.7204} & -0.0591 & 0.8074 & 0.7890
\end{tabular}
\caption{[Inpainter: ProPainter] Inpainting Quality Change on Different Input Configuration. $r_{ref}$ is a ratio of reference frames and $r_{nei}$ is a ratio of neighboring frames in input frames. $\Delta$SSIM\textsuperscript{\dag} denotes 
$\Delta$SSIM $(\times 10^{-5})$ and $\Delta$VFID\textsuperscript{\ddag} denotes $\Delta$VFID $(\times 10^{-3})$.}
\label{tab:result_ppt}
\end{table*}

The experimental results presented in Tables \ref{tab:result_ppt} and \ref{tab:result_sttn} illustrate the improvement in inpainting quality across different inpainting models and datasets. Focusing first on Table \ref{tab:result_ppt}, which reports the inpainting quality achieved with the ProPainter model, we observe that increasing the ratio of neighboring frames used as input led to substantial performance gains for videos with dynamic visual content and significant motion. However, for static video sequences, increasing the proportion of reference frames had a comparatively smaller impact on inpainting quality. An analysis of the data distribution in Figure \ref{fig:psnr_custom_metric} suggests that this discrepancy can be attributed to the scattered data points in regions with low optical flow and mask transform values. This implies that for static videos, neighboring frames already provide sufficient information to reconstruct the missing areas effectively. Given the predominance of static content in the dataset, the overall performance variation was relatively modest.

\begin{table*}[htb]
\centering
\begin{tabular}{p{10mm}|l||rrrrrr}
\multicolumn{2}{c||}{\backslashbox{\begin{tabular}[c]{@{}c@{}}Input\\ Ratio\end{tabular}}{Dataset}} & \multicolumn{3}{c|}{\textit{DAVIS}} & \multicolumn{3}{c}{\textit{MOSE}} \\ \cline{1-8} 
\multicolumn{1}{c|}{$r_{ref}$} & \multicolumn{1}{c||}{$r_{nei}$} & $\Delta$PSNR$\uparrow$ & $\Delta$SSIM\textsuperscript{\dag}$\uparrow$ & \multicolumn{1}{l|}{$\Delta$VFID\textsuperscript{\ddag}$\downarrow$} & $\Delta$PSNR$\uparrow$ & $\Delta$SSIM\textsuperscript{\dag}$\uparrow$ & $\Delta$VFID\textsuperscript{\ddag}$\downarrow$ \\ \hline\hline
0.875 & 0.125 & 0.2693 & 246.6418 & \multicolumn{1}{r|}{-15.5557} & 0.4156 & 155.6903 & -9.6891 \\
0.75  & 0.25 & 0.0249 & -22.4678 & \multicolumn{1}{r|}{5.8457} & -0.0662 & 53.4116 & -1.1583 \\
0.625 & 0.325 & 0.0444 & 41.3375 & \multicolumn{1}{r|}{-4.6894} & -0.1568 & -43.0396 & 1.2510 \\
0.5   & 0.5 & - & - & \multicolumn{1}{r|}{-} & - & - & - \\
0.325 & 0.625 & 0.1236 & 39.3755 & \multicolumn{1}{r|}{-0.1093} & 0.1063 & 26.8224 & -6.7806 \\
0.25  & 0.75 & 0.2187 & 92.9994 & \multicolumn{1}{r|}{3.6451} & 0.2717 & 68.8461 & 9.6065 \\
0.125 & 0.875 & 0.1193 & 258.3397 & \multicolumn{1}{r|}{-34.5859} & 0.0674 & 0.3274 & -16.3417 \\ \hline
\multicolumn{2}{c||}{Overall} & 0.1022 & 61.5765 & \multicolumn{1}{r|}{-2.5159} & 0.1219 & 68.3013 & -4.1592
\end{tabular}
\caption{[Inpainter: STTN] Inpainting Quality Change on Different Input Configuration. $r_{ref}$ is a ratio of reference frames and $r_{nei}$ is a ratio of neighboring frames in input frames. $\Delta$SSIM\textsuperscript{\dag} denotes 
$\Delta$SSIM $(\times 10^{-5})$ and $\Delta$VFID\textsuperscript{\ddag} denotes $\Delta$VFID $(\times 10^{-3})$.}
\label{tab:result_sttn}
\end{table*}

Turning to Table \ref{tab:result_sttn}, which presents the results with the STTN inpainting model, we find that the most pronounced performance improvements occurred in video segments characterized by rapid motion. Notably, unlike ProPainter, STTN exhibited performance gains even in the $r_{ref}=0.875$ condition, which corresponds to extremely slow motion. An intriguing observation is that the performance variations were more accurately captured by the VFID metric, which considers feature-based distances rather than pixel-based distances. This suggests that the input configuration plays a crucial role in identifying the appropriate features to fill the missing regions. This property could have significant implications for object removal tasks.

\subsubsection{Object Removal}

\begin{figure}[h]
    \centering
    \begin{tabular}{@{}c@{} @{}c@{} @{}c@{} @{}c@{}}
        $x_{comb}$ & Original & $r_{ref}>r_{nei}$ & $r_{ref}<r_{nei}$ \\
        $0.495$ 
            & \includegraphics[width=.29\linewidth]{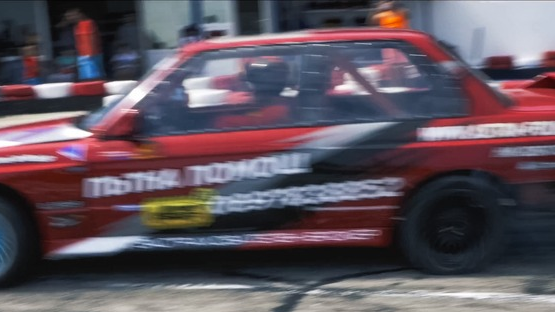} 
            & \includegraphics[width=.29\linewidth]{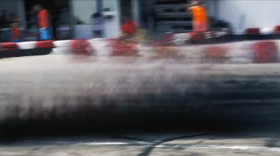} 
            & \includegraphics[width=.29\linewidth]{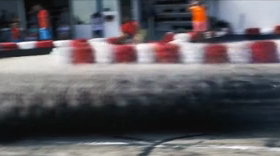} 
            \\ 
        $0.230$ 
            & \includegraphics[width=.29\linewidth]{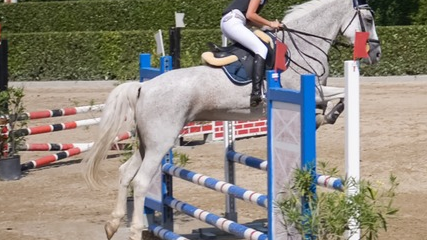} 
            & \includegraphics[width=.29\linewidth]{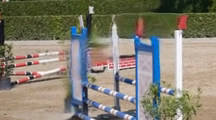} 
            & \includegraphics[width=.29\linewidth]{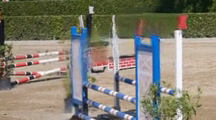} 
            \\ 
        $0.062$ 
            & \includegraphics[width=.29\linewidth]{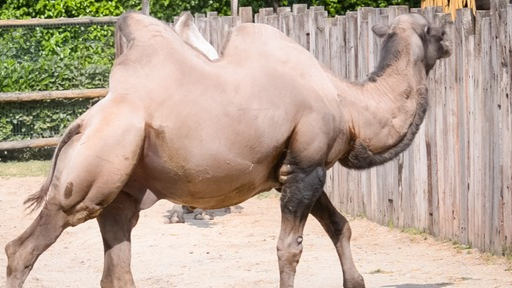} 
            & \includegraphics[width=.29\linewidth]{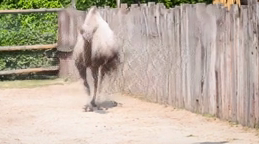} 
            & \includegraphics[width=.29\linewidth]{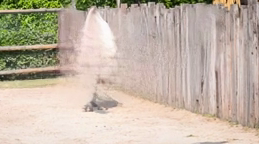} 
            \\ 
    \end{tabular}
    \caption{Results of Object Removal (Inpainter: ProPainter).}
    \label{fig:removal}
\end{figure}

The suggested framework demonstrated impressive improvement in a task for object removal. After removing objects, it shows temporal consistency on background information. In Figure \ref{fig:removal}, the results for object removal may be more intuitive to figure out the impact of {\ours}, rather than data in Table \ref{tab:result_ppt} and \ref{tab:result_sttn}. In both scenarios, whether the visual context is abrupt or gradual, {\ours} demonstrates an enhancement in the overall coherence of the video. Features obscured by objects can be retrieved from the reference frames when the scene is static and from neighboring frames when it is dynamic. This allows for the completion of frames with more detailed and temporally consistent information.

In summary, it was effective to include a dominant number of neighboring frames in the input when the visual context was rapid. Furthermore, the optical-flow-guided architecture demonstrated better qualitative improvements than those that does not adopt the flow approach.

\subsection{Memory-Quality Tradeoff}

\begin{figure}[h]
    \centering
    \begin{subfigure}[tb]{0.475\columnwidth}
        \includegraphics[width=\columnwidth]{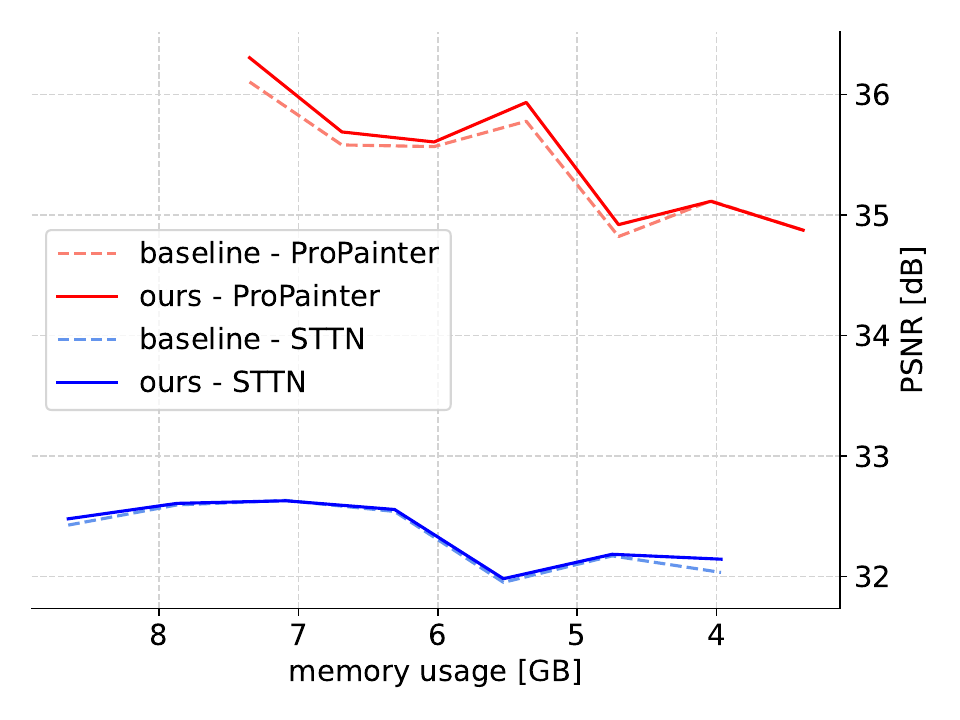}
        \caption{}
        \label{subfig:eval_memusage_psnr}
    \end{subfigure}
    \begin{subfigure}[tb]{0.475\columnwidth}
        \includegraphics[width=\columnwidth]{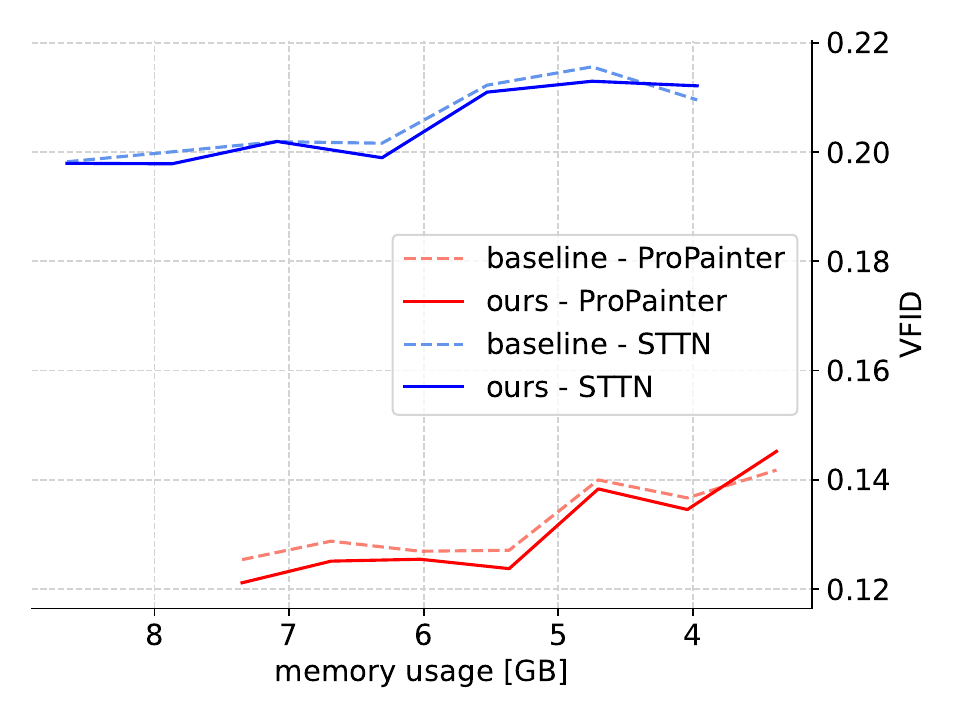}
        \caption{}
        \label{subfig:eval_memusage_vfid}
    \end{subfigure}
    \caption{Impact of Memory Reduction on Inpainting Quality. (a) represents the PSNR score which higher values are preferable and (b) depicts VFID score that is optimized at lower values. The trends illustrate the trade-off between memory efficiency and inpainting quality. As memory usage decreases, the inpainting quality also worsen, which is well demonstrated in the flow-based model, ProPainter. In the proposed input configuration, the quality improvement is more pronounced in ProPainter compared to STTN.}
    \label{fig:eval_memusage}
\end{figure}

We evaluate the degradation in inpainting quality as memory limitations were increased by progressively decreasing the number of input frames -- from 11 to 5 frames. As shown in Figure \ref{fig:eval_memusage}, as memory becomes more limited, inpainting quality decreases. However, due to our proposed input configurator, some performance improvement is maintained. With further analysis, the gap between the baseline and our framework decreases. 
When the inpainter is STTN, this tendency is relatively ambiguous, but it is clearer when it is ProPainter. This is because ProPainter performs feature propagation between neighboring frames within its networks. When the number of input frames is extremely small, performing feature propagation leads to better improvement in inpainting quality. Therefore, when memory is severely limited, the quality difference between our framework and the balanced baseline decreases.

\subsection{Influence of Distribution of Datasets}

\begin{figure}[h]
\begin{center}
\centerline{\includegraphics[width=0.6\columnwidth]{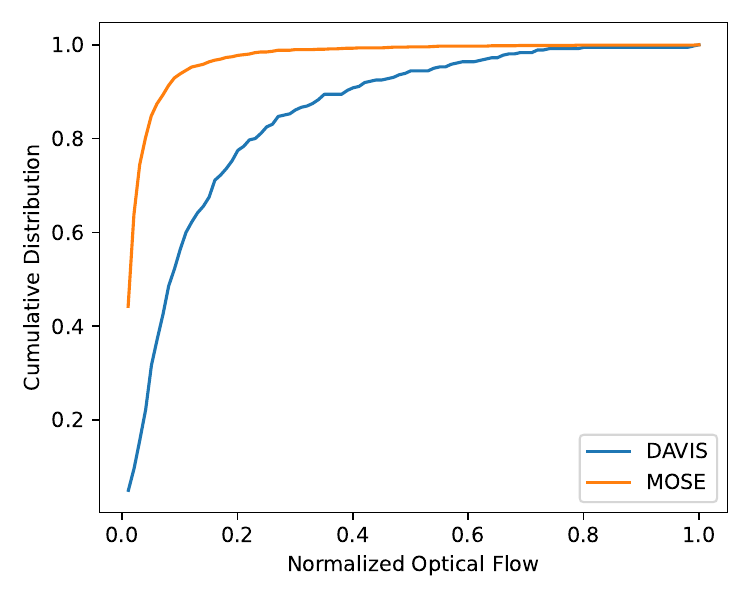}}
\caption{Flow Distribution of \textit{DAVIS} and \textit{MOSE}.}
\label{fig:data_dist}
\end{center}
\end{figure}

As mentioned in ProPainter \cite{zhou2023propainter}, the improvement is influenced by the motion distribution of the dataset. ProPainter includes image propagation and feature propagation which use flow information. {\ours} pipeline also takes advantage of optical flow. Therefore, in Table \ref{tab:result_ppt} and \ref{tab:result_sttn}, \textit{DAVIS} provides higher improvement compared to \textit{MOSE} because \textit{DAVIS} has a higher proportion of data with fast visual context, as shown in Figure \ref{fig:data_dist}.
\section{Limitations \& Future Works}

\noindent \textbf{Heuristic configuration.}
In this research, in Section \ref{subsec:configurator}, we heuristically configured input frames for improvement. Therefore, the shaded range in Figure \ref{fig:combined} can be different on different dataset on the configuration setting in Section \ref{sec:experiment}. To generalize it, it might be better to consider the knowledge of input-frame's index at a stage on model design, giving some hint for the model whether the input is the neighboring or reference frame. It may be beneficial to include a module for selecting or weighting input frames in the design.

\noindent \textbf{Data ambiguity in slow context.}
Data in the range where $r_{ref}$ is set to $[0.5, 0.875]$ is widely dispersed. As shown in Table \ref{tab:result_ppt}, not only is the improvement in quality not significant in this range, but also there is an interval where the quality even decreases.
The suggested pipeline here has space to explore methods beyond optical flow to develop more sophisticated metrics or better regression models that demonstrate clearer correlations. 

\noindent \textbf{Absence of ground truth for object removal.}
We observed that our pipeline improved inpainting quality of object removal. Unfortunately, there was not enough ways to quantify how much it improved because ground truth doesn't exist.
We attempted to measure the energy across the time in the inpainting region using the Discrete-Time Fourier Transform to measure temporal consistency on backgrounds, but we could not find significant changes in these measurement. This is because the object-removed region can be either static or dynamic, making it difficult to quantify and prove the improvement. Further research is needed to develop dataset or metrics to quantify the quality of object removal.

\noindent \textbf{Bottleneck in optical flow.}
Flow-based video inpainting models and our pipeline commonly suffer from the bottleneck on flow estimation because of high computational cost on it. Even though our approach helps to reduce computation by improving memory efficiency, the cost of flow estimation is still heavy. Fortunately, not only due to hardware advancements but also because flow estimation models are continuously being improved and proposed to be faster and better \cite{Morimitsu2024RAPIDFlow, eslami2024rethinking}, it can be expected that the performance will be increased in the future.

\begin{figure}[h]
\begin{center}
\centerline{\includegraphics[width=0.8\columnwidth]{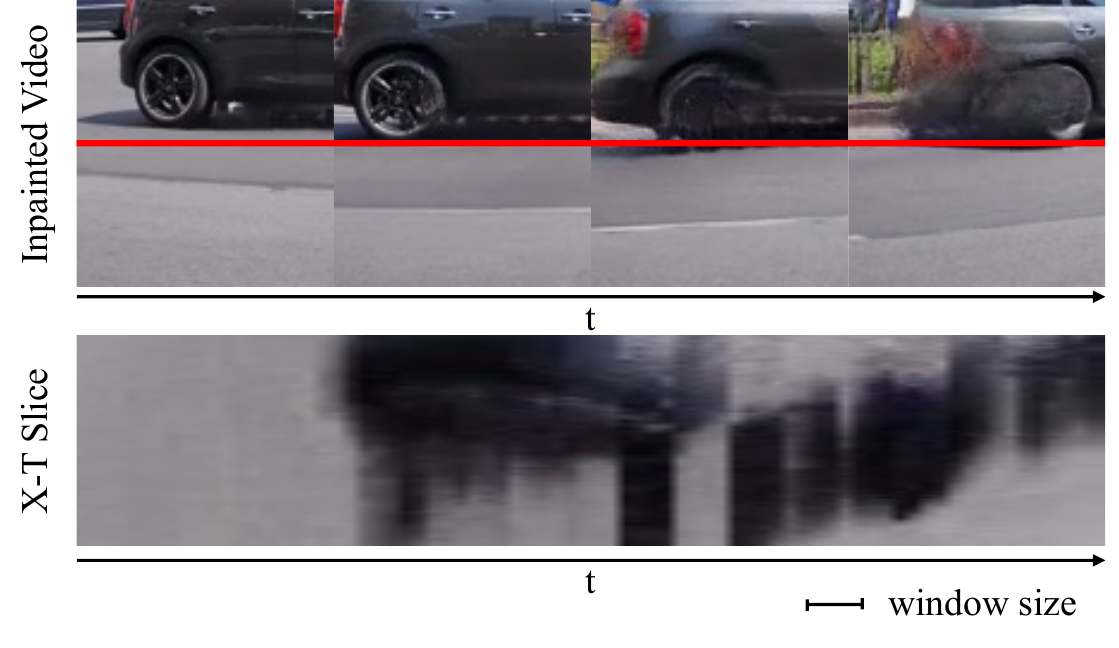}}
\caption{Temporal inconsistency with small windows. X-T slice shows the discontinuity at intervals corresponding to the window size, the number of neighboring frames.}
\label{fig:xtslice}
\end{center}
\end{figure}

\noindent \textbf{Inconsistent painting with small windows.} The inpainting models paint neighboring frames. For example, if we feed 4 neighboring frames, then they will output 4 inpainted frames. Even though the results show {\ours} can generate occluded information in more detail, this small painting window may cause temporal inconsistency, as shown in Figure \ref{fig:xtslice}. This can be improved with sliding window strategy in digital signal processing \cite{bastiaans1985sliding, lyons2012streamlining}. It provides temporal continuity by partially overlapping the windows. There are couple of studies that have improved the temporal consistency of video generation models by applying the sliding window approach \cite{bartal2024lumiere, lu2024freelong}.
\section{Conclusion}

We figured out how to compose input frames in video inpainting under memory constraints for an edge-device. When the video context is dynamic, i.e., when optical flow and mask change are significant, neighboring frames play a crucial role. Conversely, in static contexts, reference frames have a more pronounced impact on inpainting quality. 
Even the inpainter, designed to handle temporal information, were somewhat influenced by the configuration of the input. When designing video inpainting models, especially for on-device performance, the contextual flow and the affect of input frames should be considered, as discussed in this research. 

\bibliographystyle{unsrt}  
\bibliography{0_references}

\newpage
\appendix

\section*{Supplementary Materials}

\section{More Inpainting Results}
\subsection{Object Removal}

\begin{table}[h]
\centering
\begin{tabular}{m{3.42cm}|m{1cm}|m{1cm}|m{1cm}|m{3.42cm}|m{3.42cm}}
\hline 
Original frame & Optical flow & Mask change & $x_{comb}$ & Result ($r_{ref} > r_{nei}$) & Result ($r_{ref} < r_{nei}$) \\ \hline \hline

\includegraphics[width=0.22\textwidth]{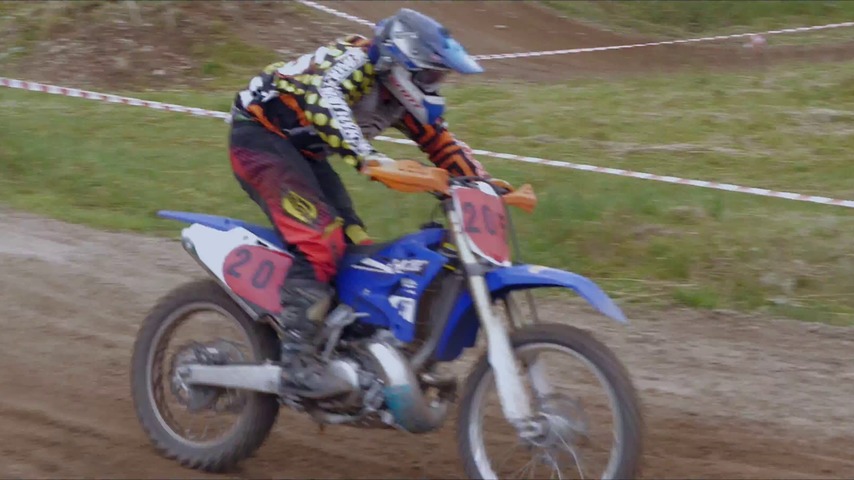} & 
0.7431 & 
0.4237 & 
0.6030 & 
\begin{tikzpicture}
    \node[anchor=south west,inner sep=0] at (0,0){\includegraphics[width=0.22\textwidth]{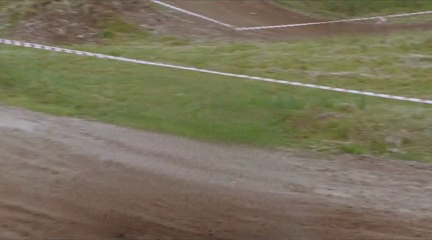}};
    \draw[red, thick] (1.2,0.4) rectangle (3.0,1.3);
\end{tikzpicture} & 
\begin{tikzpicture}
    \node[anchor=south west,inner sep=0] at (0,0){\includegraphics[width=0.22\textwidth]{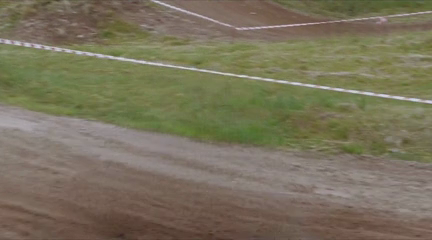}};
    \draw[red, thick] (1.2,0.4) rectangle (3.0,1.3);
\end{tikzpicture} \\

\includegraphics[width=0.22\textwidth]{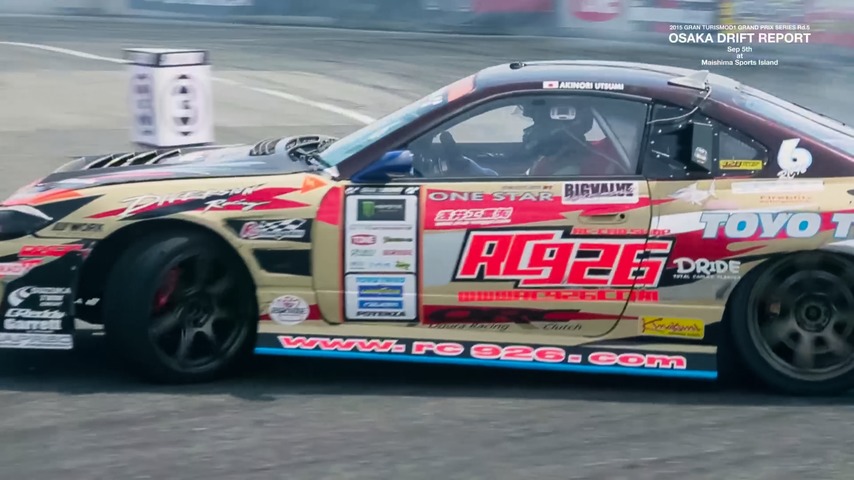} & 
0.6661 & 
0.2427 & 
0.4803 & 
\begin{tikzpicture}
    \node[anchor=south west,inner sep=0] at (0,0){\includegraphics[width=0.22\textwidth]{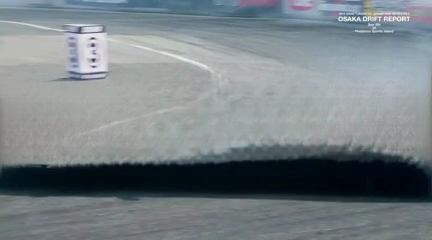}};
    \draw[red, thick] (0.1,0.6) rectangle (1.9,1.2);
\end{tikzpicture} &
\begin{tikzpicture}
    \node[anchor=south west,inner sep=0] at (0,0){\includegraphics[width=0.22\textwidth]{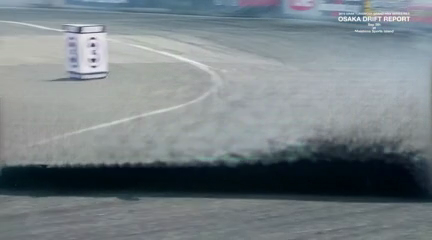}};
    \draw[red, thick] (0.1,0.6) rectangle (1.9,1.2);
\end{tikzpicture} \\

\includegraphics[width=0.22\textwidth]{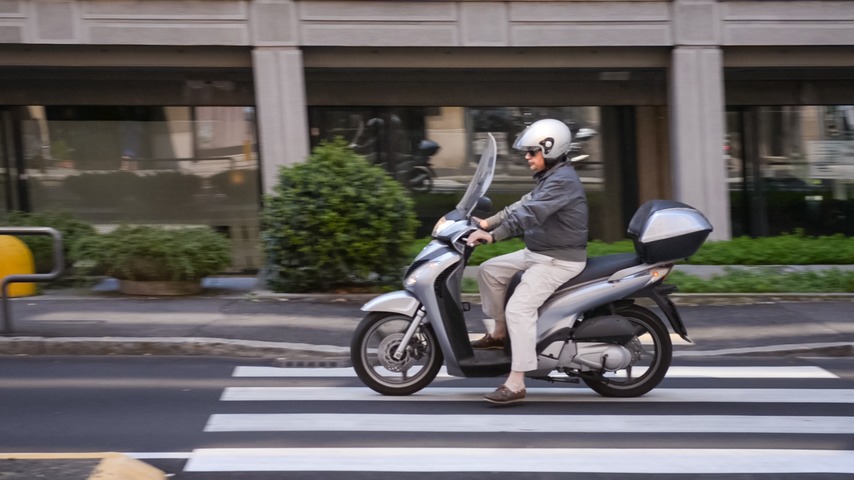} & 
0.4478 & 
0.1805 & 
0.3305 & 
\begin{tikzpicture}
    \node[anchor=south west,inner sep=0] at (0,0){\includegraphics[width=0.22\textwidth]{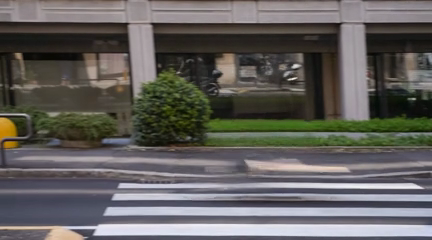}};
    \draw[red, thick] (1.7,0.3) rectangle (2.9,0.9);
\end{tikzpicture} &
\begin{tikzpicture}
    \node[anchor=south west,inner sep=0] at (0,0){\includegraphics[width=0.22\textwidth]{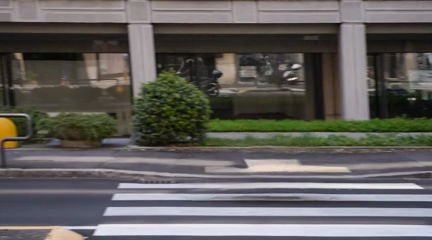}};
    \draw[red, thick] (1.7,0.3) rectangle (2.9,0.9);
\end{tikzpicture} \\

\includegraphics[width=0.22\textwidth]{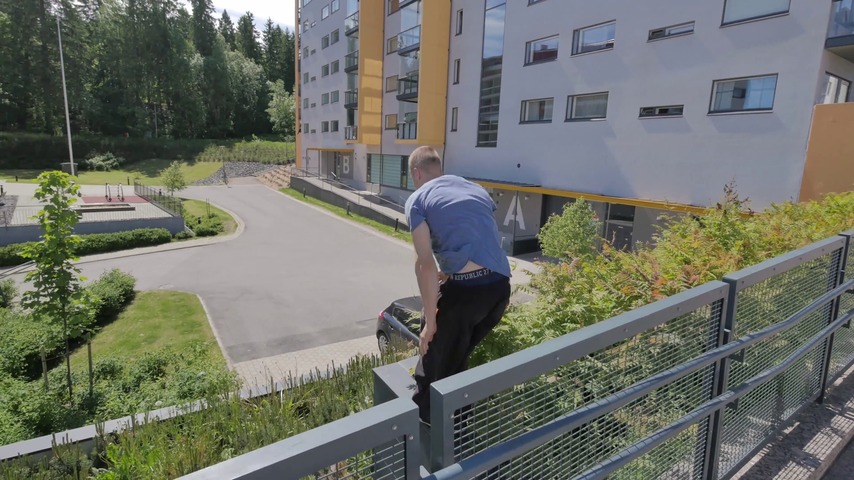} & 
0.1951 & 
0.1620 & 
0.1806 & 
\begin{tikzpicture}
    \node[anchor=south west,inner sep=0] at (0,0){\includegraphics[width=0.22\textwidth]{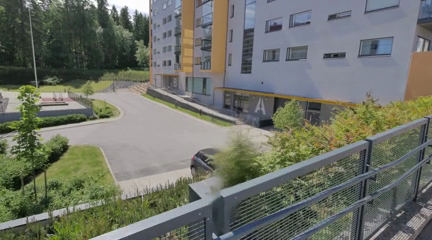}};
    \draw[red, thick] (1.65,0.15) rectangle (2.4,0.9);
\end{tikzpicture} &
\begin{tikzpicture}
    \node[anchor=south west,inner sep=0] at (0,0){\includegraphics[width=0.22\textwidth]{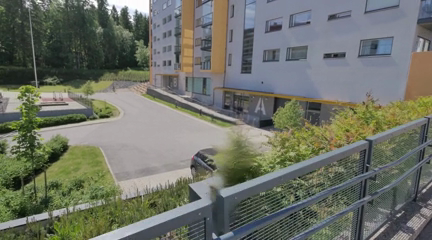}};
    \draw[red, thick] (1.65,0.15) rectangle (2.4,0.9);
\end{tikzpicture} \\

\includegraphics[width=0.22\textwidth]{Figures/appendix/object_removal/original/blackswan.jpg} & 
0.0964 & 
0.0549 & 
0.0782 & 
\begin{tikzpicture}
    \node[anchor=south west,inner sep=0] at (0,0){\includegraphics[width=0.22\textwidth]{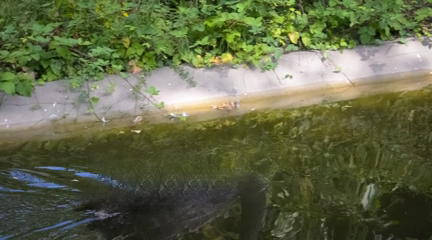}};
    \draw[red, thick] (0.85,0.25) rectangle (2.5,1.0);
\end{tikzpicture} &
\begin{tikzpicture}
    \node[anchor=south west,inner sep=0] at (0,0){\includegraphics[width=0.22\textwidth]{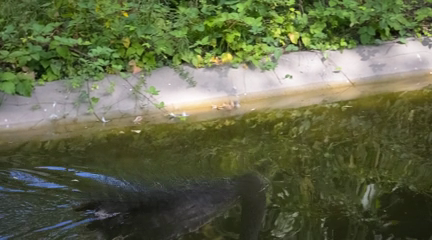}};
    \draw[red, thick] (0.85,0.25) rectangle (2.5,1.0);
\end{tikzpicture} \\

\includegraphics[width=0.22\textwidth]{Figures/appendix/object_removal/original/camel.jpg} & 
0.0531 & 
0.1102 & 
0.0781 & 
\begin{tikzpicture}
    \node[anchor=south west,inner sep=0] at (0,0){\includegraphics[width=0.22\textwidth]{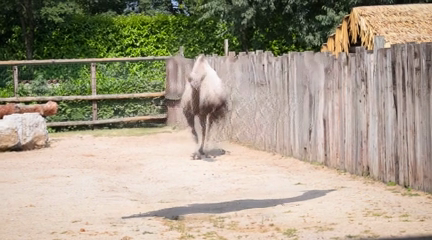}};
    \draw[red, thick] (1.3,0.4) rectangle (2.3,1.55);
\end{tikzpicture} &
\begin{tikzpicture}
    \node[anchor=south west,inner sep=0] at (0,0){\includegraphics[width=0.22\textwidth]{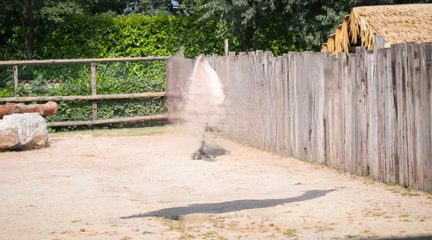}};
    \draw[red, thick] (1.3,0.4) rectangle (2.3,1.55);
\end{tikzpicture} \\


\includegraphics[width=0.22\textwidth]{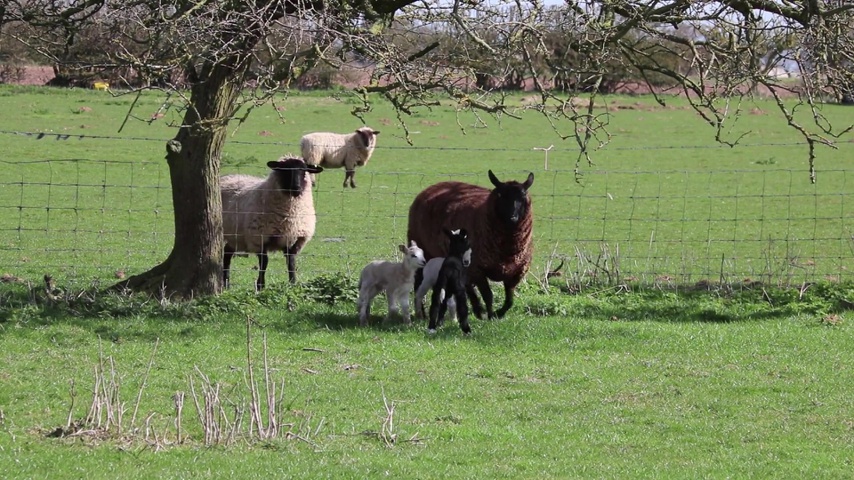} & 
0.0031 & 
0.0311 & 
0.0154 & 
\begin{tikzpicture}
    \node[anchor=south west,inner sep=0] at (0,0){\includegraphics[width=0.22\textwidth]{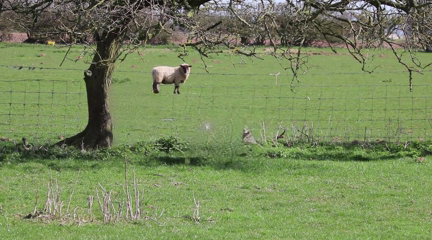}};
    \draw[red, thick] (1.6,0.7) rectangle (2.1,1.1);
\end{tikzpicture} &
\begin{tikzpicture}
    \node[anchor=south west,inner sep=0] at (0,0){\includegraphics[width=0.22\textwidth]{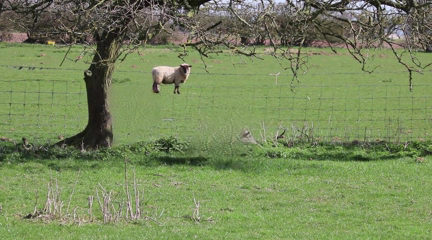}};
    \draw[red, thick] (1.6,0.7) rectangle (2.1,1.1);
\end{tikzpicture} \\ \hline

\end{tabular}
\end{table}

We fed 8 frames as inputs into the Flow-guided model, ProPainter, to get these results. The first three figures (two motorcycles and a racing car) have high dynamics in visual context. The result frames of the case where the input has more neighboring frames show more details on lawn, a line and sidewalk behind objects compared to one whose reference frames are dominant in the input set.
The fourth (a person doing Parkour) and the fifth (a black-swan) figures are cases of moderate effect of {\ours}.  In the results of the fourth one, there is more detail on a tree but less on the rail where $r_{ref} > r_{nei}$. In the results of the black-swan, the reflection on the water is more blurry on the neighboring-dominant case but the tail of water wave and natural appearance are better.
The last two images (a camel and sheep) show better background information behind the removed objects when the model was fed more reference frames in the input set.

\newpage
\subsection{Video Completion}

\begin{table}[h]
\centering
\begin{tabular}{m{3.8cm}|m{1.0cm}|m{1.0cm}|m{1.0cm}|m{3.8cm}|m{3.8cm}}
\hline 
Original frame & Optical flow & Mask change & $x_{comb}$ & Result ($r_{ref} > r_{nei}$) & Result ($r_{ref} < r_{nei}$) \\ \hline \hline

\includegraphics[width=0.22\textwidth]{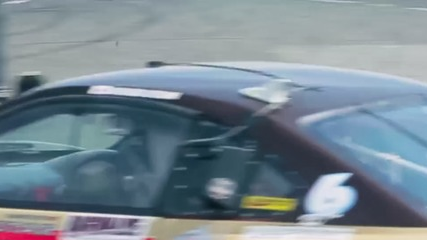} & 
0.7928 & 
0.0876 & 
0.4835 & 
\begin{tikzpicture}
    \node[anchor=south west,inner sep=0] at (0,0){\includegraphics[width=0.22\textwidth]{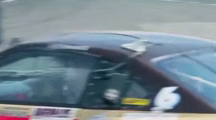}};
    \draw[red, thick] (0,0.7) rectangle (2,1.6);
\end{tikzpicture} & 
\begin{tikzpicture}
    \node[anchor=south west,inner sep=0] at (0,0){\includegraphics[width=0.22\textwidth]{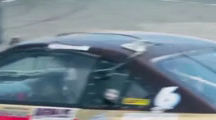}};
    \draw[red, thick] (0,0.7) rectangle (2,1.6);
\end{tikzpicture} \\

\includegraphics[width=0.22\textwidth]{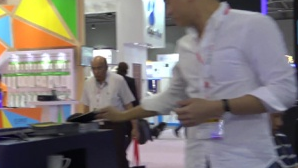} & 
0.4634 & 
0.1335 & 
0.3187 & 
\begin{tikzpicture}
    \node[anchor=south west,inner sep=0] at (0,0){\includegraphics[width=0.22\textwidth]{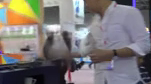}};
    \draw[red, thick] (0.9,0.5) rectangle (2.1,1.8);
\end{tikzpicture} &
\begin{tikzpicture}
    \node[anchor=south west,inner sep=0] at (0,0){\includegraphics[width=0.22\textwidth]{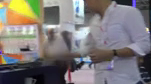}};
    \draw[red, thick] (0.9,0.5) rectangle (2.1,1.8);
\end{tikzpicture} \\

\includegraphics[width=0.22\textwidth]{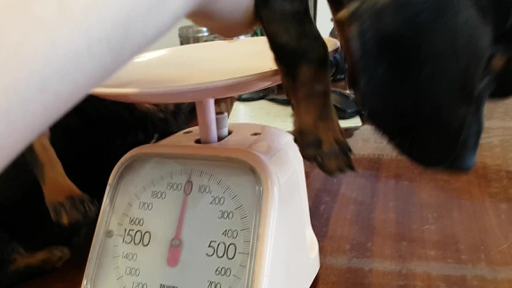} & 
0.2078 & 
0.1873 & 
0.1988 & 
\begin{tikzpicture}
    \node[anchor=south west,inner sep=0] at (0,0){\includegraphics[width=0.22\textwidth]{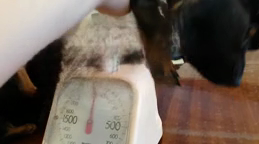}};
    \draw[red, thick] (0.7,0.7) rectangle (2.2,2.1);
\end{tikzpicture} &
\begin{tikzpicture}
    \node[anchor=south west,inner sep=0] at (0,0){\includegraphics[width=0.22\textwidth]{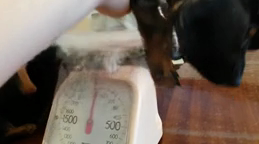}};
    \draw[red, thick] (0.7,0.7) rectangle (2.2,2.1);
\end{tikzpicture} \\

\includegraphics[width=0.22\textwidth]{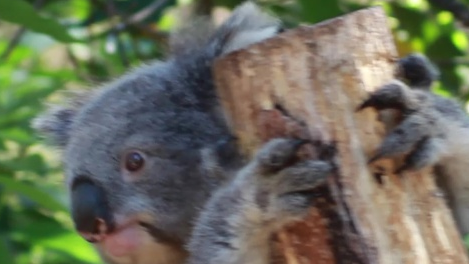} & 
0.1115 & 
0.1211 & 
0.1157 & 
\begin{tikzpicture}
    \node[anchor=south west,inner sep=0] at (0,0){\includegraphics[width=0.22\textwidth]{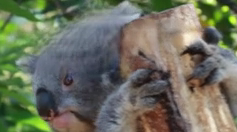}};
    \draw[red, thick] (0.7,1.1) rectangle (2.3,2.1);
\end{tikzpicture} &
\begin{tikzpicture}
    \node[anchor=south west,inner sep=0] at (0,0){\includegraphics[width=0.22\textwidth]{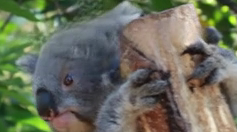}};
    \draw[red, thick] (0.7,1.1) rectangle (2.3,2.1);
\end{tikzpicture} \\

\includegraphics[width=0.22\textwidth]{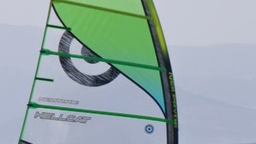} & 
0.0877 & 
0.0911 & 
0.0892 & 
\begin{tikzpicture}
    \node[anchor=south west,inner sep=0] at (0,0){\includegraphics[width=0.22\textwidth]{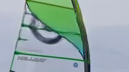}};
    \draw[red, thick] (0.5,1.1) rectangle (2.2,1.8);
\end{tikzpicture} &
\begin{tikzpicture}
    \node[anchor=south west,inner sep=0] at (0,0){\includegraphics[width=0.22\textwidth]{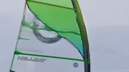}};
    \draw[red, thick] (0.5,1.1) rectangle (2.2,1.8);
\end{tikzpicture} \\

\includegraphics[width=0.22\textwidth]{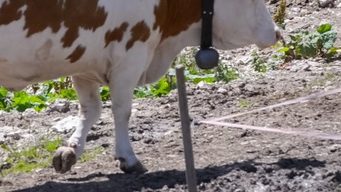} & 
0.0491 & 
0.1422 & 
0.0899 & 
\begin{tikzpicture}
    \node[anchor=south west,inner sep=0] at (0,0){\includegraphics[width=0.22\textwidth]{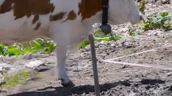}};
    \draw[red, thick] (0.3,0.2) rectangle (1.2,1.8);
\end{tikzpicture} &
\begin{tikzpicture}
    \node[anchor=south west,inner sep=0] at (0,0){\includegraphics[width=0.22\textwidth]{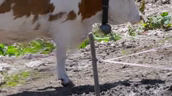}};
    \draw[red, thick] (0.3,0.2) rectangle (1.2,1.8);
\end{tikzpicture} \\

\includegraphics[width=0.22\textwidth]{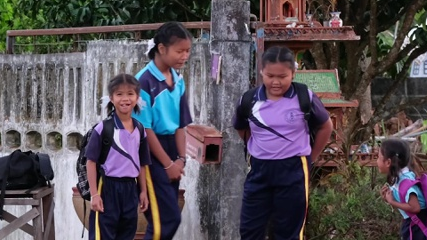} & 
0.0064 & 
0.1556 & 
0.0718 & 
\begin{tikzpicture}
    \node[anchor=south west,inner sep=0] at (0,0){\includegraphics[width=0.22\textwidth]{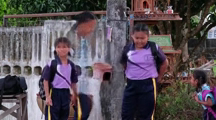}};
    \draw[red, thick] (1.2,0.2) rectangle (1.9,1.5);
\end{tikzpicture} &
\begin{tikzpicture}
    \node[anchor=south west,inner sep=0] at (0,0){\includegraphics[width=0.22\textwidth]{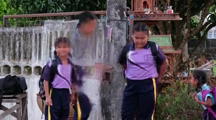}};
    \draw[red, thick] (1.2,0.2) rectangle (1.9,1.5);
\end{tikzpicture} \\ \hline

\end{tabular}
\end{table}

As discussed in Sections \ref{sec:observation} and \ref{sec:experiment}, when visual dynamics is high (first three images), neighboring frames have more influence on quality improvement.

In relatively static scenes (last four images), as the data points are dispersed in Figure \ref{subfig:pro_dyn_flow} and the quality differences are not significant in Table \ref{tab:result_ppt}, there were cases where neighboring frames resulted in better results, while in other cases, reference frames yielded better outcomes.

\end{document}